\documentclass[10pt,twocolumn,letterpaper]{article}

\PassOptionsToPackage{table,xcdraw}{xcolor}
\usepackage[pagenumbers]{cvpr}

\usepackage[T1]{fontenc}
\usepackage{graphicx}
\usepackage{booktabs}
\usepackage{multirow}
\usepackage{amsmath}
\usepackage{amssymb}
\usepackage{xcolor}
\usepackage[normalem]{ulem}
\useunder{\uline}{\ul}{}
\usepackage{pifont}
\usepackage{makecell}
\newcommand{\cmark}{\ding{51}}

\usepackage{tikz}

\definecolor{blip2color}{rgb}{0.961,0.871,0.702}
\definecolor{qwen3color}{rgb}{0.694,0.886,0.988}

\DeclareRobustCommand{\blipdot}{\raisebox{-0.65pt}{%
\begin{tikzpicture}
\fill[blip2color] (0,0) circle (.8ex);
\draw[black, line width=0.5pt] (0,0) circle (.8ex);
\end{tikzpicture}}}

\DeclareRobustCommand{\qwendot}{\raisebox{-0.65pt}{%
\begin{tikzpicture}
\fill[qwen3color] (0,0) circle (.8ex);
\draw[black, line width=0.5pt] (0,0) circle (.8ex);
\end{tikzpicture}}}

\newcommand{\paragraphbf}[1]{\vspace{0.35em}\noindent\textbf{#1}\quad\ignorespaces}

\definecolor{cvprblue}{rgb}{0.21,0.49,0.74}
\usepackage[breaklinks,colorlinks,allcolors=cvprblue]{hyperref}

\title{Matching in Context: Temporal Fusion and Query Adaptation for Training-Free Video Temporal Grounding}

\author{Zhuo Cao$^1$, Bingqing Zhang$^1$, Sen Wang$^1$, Xue Li$^1$\thanks{Corresponding author.}\\
$^1$The University of Queensland, Australia\\
{\tt\small \{william.cao, bingqing.zhang, sen.wang\}@uq.edu.au}\\
{\tt\small xueli@eesc.uq.edu.au}
}

\hypersetup{
    pdftitle={Matching in Context: Temporal Fusion and Query Adaptation for Training-Free Video Temporal Grounding},
    pdfauthor={Zhuo Cao, Bingqing Zhang, Sen Wang, Xue Li}
}

\begin{document}
\maketitle
\begin{abstract}
  Text-guided Video Temporal Grounding (VTG) localizes moments in an untrimmed video that correspond to a text query. Supervised approaches require costly temporal annotations and task-specific training, yet can generalize poorly under distribution shift. To reduce this dependence on task-specific supervision, training-free approaches use pretrained vision-language models to localize moments by measuring the similarity between video and query features. However, independently encoded frames provide limited temporal context, while a fixed query embedding can assign similar scores to different candidate moments within the video. To address these limitations, we propose DSE-VTG, a \underline{D}ual-\underline{S}ide \underline{E}nhancement framework that combines visual evidence across temporal granularities with video-specific query adaptation, without task-specific training. On the visual side, Multi-scale Similarity Fusion (MSF) combines frame- and clip-level similarities into a unified, temporally aware similarity profile. On the query side, Query-level Test-Time Adaptation (Q-TTA) uses pseudo-labels to optimize a lightweight additive offset for more discriminative matching within the video, without fine-tuning the backbone or calling external large language models. Experiments on three standard and two out-of-distribution (OOD) benchmarks show gains over the prior training-free methods. On Charades-STA, DSE-VTG improves mIoU by 5.89 percentage points over the strongest compared baseline under the same backbone. Under distribution shift, DSE-VTG reaches 50.86 mIoU on Charades-CG Novel-Word, surpassing the strongest supervised baseline by 2.76 mIoU. Our code will be released upon acceptance.
\end{abstract}

\section{Introduction}
\label{sec:intro}
As the primary visual medium in daily life, video contains a density of spatiotemporal information that far exceeds that of static images. The complexity of video streams necessitates intelligent systems capable of accurately parsing dynamic visual information. As a crucial step toward this goal, text-guided Video Temporal Grounding (VTG) aims to localize temporal segments in untrimmed videos that semantically correspond to natural language queries. This task serves as a cornerstone for content production and consumption industries, including video editing, summarization, and question answering.

\begin{figure*}[t]
    \centering
    \begin{minipage}{0.65\textwidth}
        \centering
        \includegraphics[width=0.95\linewidth]{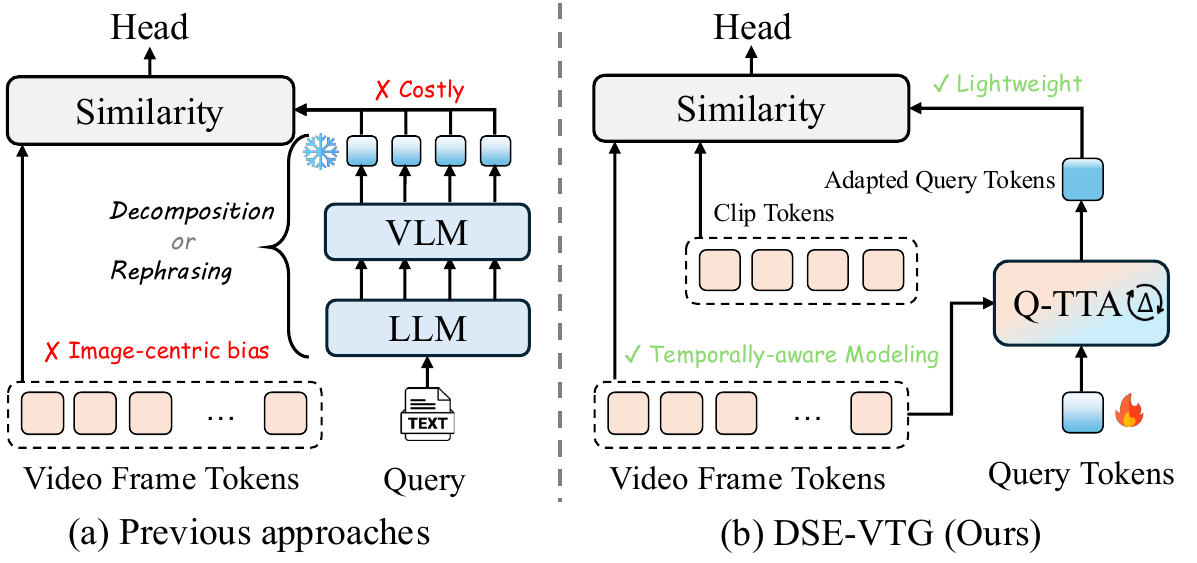}
        \caption{Different architectures for training-free video temporal grounding.}
        \label{fig:teaser_fig1}
    \end{minipage}\hfill
    \begin{minipage}{0.35\textwidth}
        \centering
        \includegraphics[width=\linewidth]{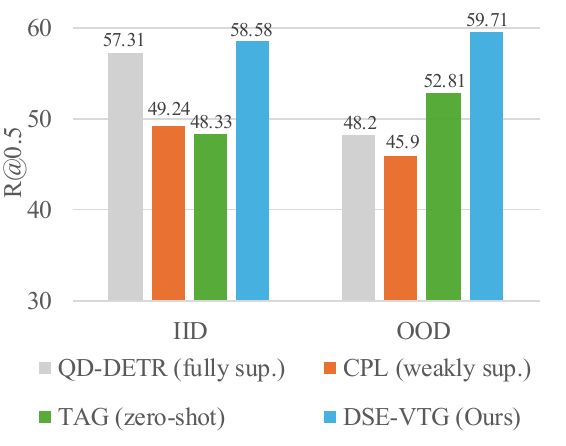}
        \caption{IID and OOD performance comparison on Charades-STA/CG~\mbox{\cite{gao2017tall, li2022charadesCG}}.}
        \label{fig:teaser_fig2}
    \end{minipage}
\end{figure*}

Despite the broad utility of this task, achieving precise temporal localization remains challenging, owing to the difficulty of establishing fine-grained semantic alignment between unconstrained visual sequences and diverse linguistic queries. Historically, fully supervised methods have dominated this field~\cite{lei2021detecting, moon2023qddetr, liu2024tuning, Cao_2025_WACV} and have achieved promising results. However, reliance on expensive temporal annotations and susceptibility to out-of-distribution (OOD) shifts have severely limited the scalability of these approaches. Training-free methods reduce this dependence on task-specific supervision~\cite{luo2024zero, Zheng2024tfvtg, lee2025tag, xu2024vtg}. By directly leveraging the rich open-vocabulary representations of pretrained Vision-Language Models (VLMs), such as BLIP-2~\cite{li2023blip}, these methods perform temporal grounding through inference-time cross-modal matching. This matching strategy enables temporal grounding without task-specific training or temporal annotations.

For similarity-based training-free VTG~\cite{luo2024zero, Zheng2024tfvtg, lee2025tag}, grounding depends on both the temporal evidence represented by the visual features and the ability of query-video matching to distinguish candidate moments. On the visual side, independently encoded frames provide fine temporal resolution but do not directly encode how an event unfolds across consecutive frames. To exploit temporal structure, TFVTG~\cite{Zheng2024tfvtg} evaluates event proposals through changes in frame-query similarity and contrasts between candidate segments and their surroundings. TAG~\cite{lee2025tag} aggregates neighboring frame features through temporal pooling and forms proposals using temporal-coherence clustering (TCC). These strategies account for temporal relationships when constructing and scoring candidate segments. However, their underlying visual features are still extracted independently, without jointly encoding visual changes across consecutive frames. On the query side, moments that share objects or actions with a query may receive high similarity scores even when they only partially match the described event. These partial matches can make it difficult to distinguish the target moment using a fixed query embedding. To improve query-video matching, prior works parse queries into semantic components~\cite{luo2024zero} or use external LLMs for query decomposition and rephrasing~\cite{Zheng2024tfvtg, xu2024vtg} (as shown in Fig.~\mbox{\ref{fig:teaser_fig1}}(a)). These strategies make the query content more explicit. However, the resulting query embeddings remain fixed during matching, without adaptation to video-specific visual evidence. And the use of external LLMs also incurs additional inference overhead.

To address these challenges, we introduce DSE-VTG (illustrated in Fig.~\ref{fig:teaser_fig1}(b)), a dual-side enhancement framework for training-free temporal grounding. The framework combines evidence across temporal granularities through MSF and adapts query-video matching through Q-TTA. On the video side, we introduce a Multi-scale Similarity Fusion (MSF) module, which combines temporal evidence at two granularities without introducing any trainable visual parameters. By utilizing a backbone capable of video processing, we obtain both fine-grained frame-level features and temporally aware clip-level features. The similarity scores of these features are subsequently projected onto a unified timeline via center-projection alignment. This fusion combines fine temporal resolution with the broader context encoded by clips. On the query side, Query-level Test-Time Adaptation (Q-TTA) optimizes a lightweight additive offset to the query embedding during inference. Pseudo-labels derived from the initial frame-query similarities guide this update without changing any VLM parameters. The adapted query embedding is used to compute both frame-level and clip-level matching scores. MSF then combines these scores for temporal grounding.

Both visual streams are independent of the query, so their embeddings can be extracted offline and reused across queries. Each new query requires text encoding, lightweight adaptation, and matching against the same cached video features. This retains the multiple-query reuse supported by feature-based training-free pipelines.

To validate the effectiveness of the proposed method, we conduct evaluations across three standard benchmarks, Charades-STA~\cite{gao2017tall}, ActivityNet Captions~\cite{krishna2017dense}, QVHighlights~\cite{lei2021detecting}, and two out-of-distribution benchmarks derived from Charades-STA, \ie, Charades-CG~\cite{li2022charadesCG} and Charades-CD~\cite{charadesCD}. DSE-VTG consistently improves over strong training-free baselines across the evaluated benchmarks. As illustrated by the IID comparison in Fig.~\ref{fig:teaser_fig2}, it improves R@0.5 over the same-backbone TAG baseline~\cite{lee2025tag} by 10.25 percentage points on Charades-STA. On QVHighlights, it also improves both localization recall and mean average precision. Notably, under the out-of-distribution setting, DSE-VTG maintains strong robustness. As shown in Fig.~\ref{fig:teaser_fig2}, it achieves 59.71 R@0.5 on the Charades-CG Novel-Word split, even outperforming the supervised baselines, and consistently improves performance on Charades-CD. Further analysis demonstrates that the proposed method generalizes well across multiple backbones. These results confirm that DSE-VTG effectively strengthens the temporal semantic alignment and generality of the training-free paradigm.

Overall, the main contributions of this work are summarized as follows: 

(1) We propose DSE-VTG, a dual-side enhancement framework for training-free video temporal grounding that combines temporal context beyond individual frames with query adaptation for more discriminative matching within a video. 

(2) We design two specific functional components: the Multi-scale Similarity Fusion (MSF) module for unified temporally aware visual modeling, and the Query-level Test-Time Adaptation (Q-TTA) module for lightweight, dynamic semantic alignment.

(3) Experiments on three standard and two OOD benchmarks demonstrate gains over strong training-free baselines. The same-backbone comparisons further show improvements across different pretrained representations.
\section{Related Work}
\label{sec:related}
\subsection{Video Temporal Grounding}
Video temporal grounding (VTG) has been studied under fully supervised~\cite{lei2021detecting, moon2023qddetr, lin2023univtg,liu2024tuning, Cao_2025_WACV, Cao2025flashmmr}, weakly supervised~\cite{duan2018weakly, zheng2022weakly_cvpr, huang2023weakly}, and unsupervised paradigms~\cite{KPSC,nam2021zero, wang2022prompt, zheng2023generating}. To reduce the need for task-specific supervision and optimization, training-free methods~\cite{luo2024zero} instead use pretrained Vision-Language Models (VLMs)~\cite{radford2021clip,li2023blip} to match queries with video content directly at inference time. TFVTG~\cite{Zheng2024tfvtg} models temporal changes in frame-query similarities and scores proposals using their internal and surrounding context. To address semantic fragmentation and skewed similarity distributions, TAG~\cite{lee2025tag} introduces temporal pooling and temporal-coherence clustering.

Recent zero-shot methods also use language generation and multimodal reasoning. DART~\cite{dart2026} selects diverse, query-relevant keyframes and routes queries between direct prediction and structured temporal reasoning according to an estimated difficulty. GranAlign~\cite{jeon2026granalign} pairs semantically coarse and fine query rewrites with query-agnostic and query-aware video captions to reduce granularity mismatch. These methods enrich the descriptions or reasoning used for localization, complementing approaches that aggregate frame-query similarity evidence along the video timeline.

For the frame-feature matching pipelines of Luo et al.~\cite{luo2024zero}, TFVTG~\cite{Zheng2024tfvtg}, and TAG~\cite{lee2025tag}, temporal processing acts on features extracted from individual frames. Meanwhile, recent video-language foundation models~\cite{wang2022internvideo,li2026qwen3vl-embed} can provide temporally aware representations across multiple granularities, from individual frames to multi-second clips. However, these capabilities remain under-exploited in training-free VTG, where actions and events span multiple temporal scales. In contrast to single-resolution approaches, our method combines frame- and clip-level evidence for training-free temporal grounding.

\subsection{Query Understanding and Semantic Alignment}
Query understanding methods make the semantic content of a query more explicit for retrieval. In training-free VTG, TFVTG~\cite{Zheng2024tfvtg} decomposes queries into sub-events, while VTG-GPT~\cite{xu2024vtg} rewrites queries and matches them against generated video captions. In conversational search, AdaRewriter~\cite{Lai2025AdaRewriter} uses a learned reward model to select query reformulations at inference time. These methods refine query text through rewriting or the selection of candidate reformulations. Their adaptation operates at the textual level rather than through direct optimization of the query embedding used for matching.

Test-time adaptation also uses visual evidence to adjust alignment. For zero-shot temporal action localization, T3AL~\cite{Liberatori_2024t3al} infers a video-level action pseudo-label, adapts a pretrained VLM to the test video, and refines action proposals with frame-level captions. Its adaptation signal comes from the test video rather than a rewritten query, and it updates the representation model used for localization. T3AL targets action categories, whereas VTG requires distinguishing the moment described by a free-form query. In this query-conditioned setting, a semantically clear description can still match several competing moments. Our Q-TTA addresses this matching problem by adapting the query representation to the current video while keeping the backbone fixed.

\section{Method}
\label{sec:method}

\subsection{Problem Formulation}
We study the task of Video Temporal Grounding (VTG) under a zero-shot protocol which is free of task-specific training: no component is trained or fine-tuned on any VTG dataset, and the only test-time optimization is a disposable per-query offset. Given an untrimmed video $\mathcal{V}$ and a natural language query $\mathcal{Q}$, VTG aims to predict one or more temporal moments $\{(t_s^{k}, t_e^{k}, c^{k})\}_{k=1}^{K}$ in $\mathcal{V}$ that semantically correspond to $\mathcal{Q}$. Here $t_s^{k}$ and $t_e^{k}$ are the start and end timestamps of the $k$-th predicted moment, and $c^{k}$ denotes its branch-normalized proposal score.

\begin{figure*}[t]
\centering
\includegraphics[width=\linewidth]{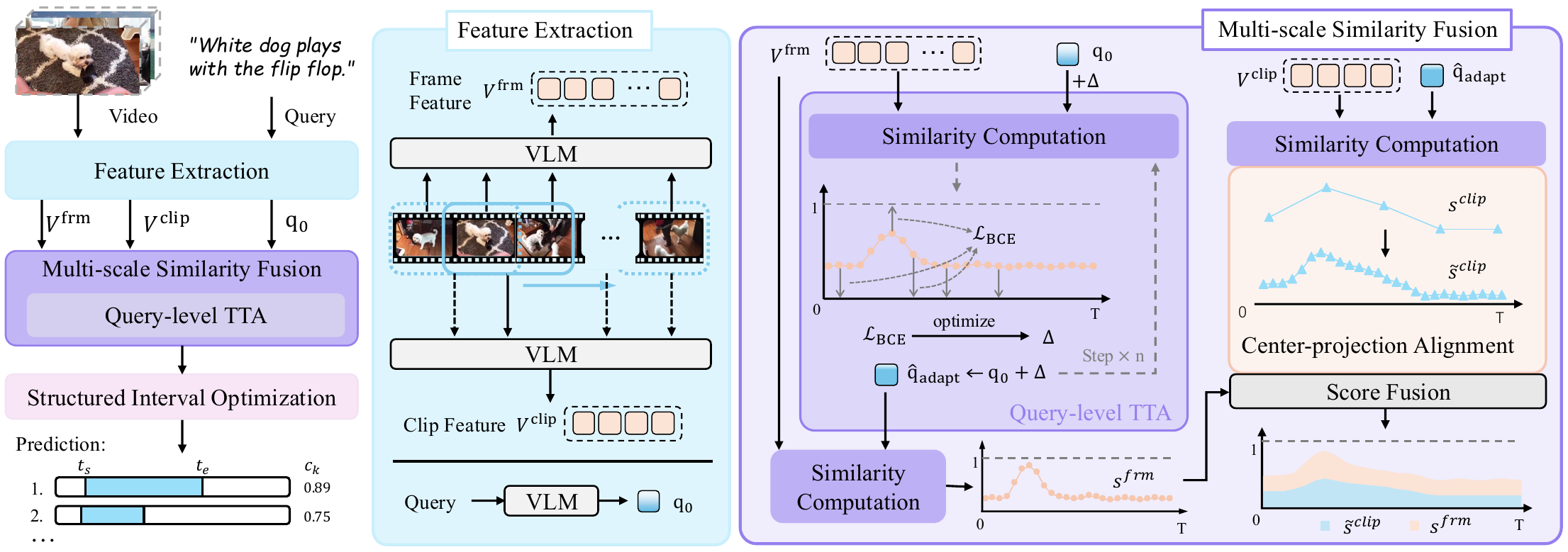}
\caption{\textbf{Overall architecture of DSE-VTG.} A VLM extracts frame- and clip-level visual features $V^{\mathrm{frm}}$ and $V^{\mathrm{clip}}$, together with the initial query embedding $\mathbf{q}_0$. The Query-level TTA module adapts the query to obtain $\hat{\mathbf{q}}_{\mathrm{adapt}}$. Multi-scale Similarity Fusion computes frame- and clip-level similarities, projects clip scores onto the frame timeline through Center-projection Alignment, and fuses the two streams. Structured interval optimization decodes the fused scores into temporal boundaries.}
\vspace{-1mm}
\label{fig:model}
\end{figure*}

\subsection{Overview}
Figure~\ref{fig:model} shows the overall architecture of DSE-VTG. A video-capable VLM encodes the video into frame features $V^{\text{frm}}$ and clip features $V^{\text{clip}}$, and the query into an embedding $\mathbf{q}_0$ (Sec.~\ref{subsec:feature extraction}). Q-TTA adapts the query embedding using pseudo-labels derived from the initial frame--query similarities (Sec.~\ref{subsec:q_tta}). MSF projects the two similarity streams into a common frame timeline (Sec.~\mbox{\ref{subsec:msf}}), and structured interval optimization decodes the fused scores into temporal boundaries (Sec.~\mbox{\ref{subsec:loc_head}}). The visual embeddings are independent of the query and can be computed offline once per video. Each new query reuses these caches for adaptation and localization.

\subsection{Multi-Granularity Feature Extraction}
\label{subsec:feature extraction}
To establish a robust foundation for temporal grounding, we first map the raw video $\mathcal{V}$ and text query $\mathcal{Q}$ into a shared semantic space. Existing training-free pipelines typically inherit an image-centric representation bias from image-pretrained vision-language models, treating videos as independent frames. To inject temporal cues without additional training, we adopt a video-capable backbone to extract multi-granularity visual features, capturing both fine-grained local semantics and broader temporal dynamics.

\paragraphbf{Video-capable backbone}
Unlike image-only architectures (\eg, ViT + Q-Former~\cite{li2023blip} and CLIP-style encoders~\cite{radford2021clip}), a video-capable backbone natively supports multi-frame inputs and models temporal relations during feature extraction. This yields temporally aware clip representations while preserving training-free inference.

\paragraphbf{Query and dual-granularity video features}
Given a natural language query $\mathcal{Q}$, we first extract its sentence-level text embedding $\mathbf{q}_0 \in \mathbb{R}^D$, where $D$ is the feature dimension. For the visual counterpart, let a video be represented as a sequence of $T$ sampled frames. We extract two complementary feature streams: (i) \textbf{Frame-level features} sampled at 3\,fps by default, and (ii) \textbf{Clip-level features} obtained by applying a sliding temporal window of length $W$ seconds with stride $S$ seconds. To better cover temporal boundaries, we also include boundary windows of length $B=W/2$. The resulting $N$ clip windows include these boundary windows. The extracted frame- and clip-level features are respectively denoted as:
\begin{equation}
V^{\text{frm}} = \{\mathbf{v}^{\text{frm}}_{t}\}_{t=1}^{T},\;\; V^{\text{clip}} = \{\mathbf{v}^{\text{clip}}_{i}\}_{i=1}^{N},
\end{equation}
where $\mathbf{v}^{\mathrm{frm}}_t,\mathbf{v}^{\mathrm{clip}}_i\in\mathbb{R}^D$.

\subsection{Query-level Test-Time Adaptation}
\label{subsec:q_tta}
A fixed query embedding can score competing moments similarly even when the text is clear. Q-TTA uses the initial frame-level matches as video-specific supervision to adapt the query embedding to the current video. Unlike common TTA methods that adapt visual encoders or task-specific heads, Q-TTA optimizes a single additive offset to the query embedding while keeping all VLM parameters fixed. The offset is initialized separately for each video--query pair and discarded after inference.

\paragraphbf{Pseudo-label construction}
To establish the baseline alignment for pseudo-labeling, we first compute the cosine similarity between the initial text embedding $\mathbf{q}_0$ and frame-level visual features $V^{\text{frm}}$. Denoting $\ell_2$-normalized features by $\hat{\cdot}$, the initial frame-level similarity is computed as:
\begin{equation}
\label{eq:init_sim}
s^{(0)}_t = \langle \hat{\mathbf{q}}_0, \hat{\mathbf{v}}^{\text{frm}}_{t} \rangle, \quad t=1,\dots,T.
\end{equation}
For $T\geq2$, we select the highest-scoring $k_{\mathrm{pos}}$ frames as pseudo-positives $\mathcal{P}$ and sample $k_{\mathrm{neg}}$ distinct pseudo-negatives uniformly from the remaining frames to form $\mathcal{N}$. Let $\gamma_{\mathrm{pos}}$ and $\gamma_{\mathrm{neg}}$ denote the positive and negative sampling ratios, respectively. The sample counts are
{\begin{equation}
\begin{aligned}
k_{\mathrm{pos}}&=\min\{\max\{1,\operatorname{round}(\gamma_{\mathrm{pos}}T)\},T-1\},\\
k_{\mathrm{neg}}&=\min\{\max\{1,\operatorname{round}(\gamma_{\mathrm{neg}}T)\},T-k_{\mathrm{pos}}\}.
\end{aligned}
\end{equation}}
The sets are fixed throughout adaptation. Adaptation is skipped when fewer than two frames are available.

\paragraphbf{Adaptation objective}
We introduce a learnable offset vector $\Delta \in \mathbb{R}^D$ to adjust the query embedding. The adapted and $\ell_2$ normalized query is defined as:
{\begin{equation}
\hat{\mathbf{q}}_{\mathrm{adapt}}=\frac{\mathbf{q}_0+\Delta}{\|\mathbf{q}_0+\Delta\|_2+\epsilon}.
\end{equation}}
At each adaptation step, we recompute the frame similarities under the adapted query and rescale them to $[0,1]$ to obtain bounded logits:
\begin{equation}
l_t = \langle \hat{\mathbf{q}}_{\text{adapt}}, \hat{\mathbf{v}}^{\text{frm}}_{t} \rangle,
\quad
\tilde{l}_t = \frac{l_t - \min(\mathbf{l})}{\max(\mathbf{l}) - \min(\mathbf{l}) + \epsilon},
\end{equation}
where $\mathbf{l} = \{l_t\}^{T}_{t=1}$ collects the per-frame similarities and $\epsilon$ is a small constant for numerical stability.

The offset $\Delta$ is optimized by minimizing a binary cross-entropy objective supervised by the pseudo-labels, where $\mathcal{L}_{\text{BCE}}(\tilde{l}_t, y)$ denotes the logits formulation, i.e., a sigmoid is applied to $\tilde{l}_t$ inside the loss:
\begin{equation}
\mathcal{L}(\Delta) = \frac{w_{\text{pos}}}{|\mathcal{P}|} \sum_{t \in \mathcal{P}} \mathcal{L}_{\text{BCE}}(\tilde{l}_t, 1) + \frac{w_{\text{neg}}}{|\mathcal{N}|} \sum_{t \in \mathcal{N}} \mathcal{L}_{\text{BCE}}(\tilde{l}_t, 0),
\end{equation}
where \mbox{$w_{\text{pos}}$} and \mbox{$w_{\text{neg}}$} are loss weights for positive and negative frames, respectively. 
The objective encourages higher normalized scores for pseudo-positive frames and lower scores for pseudo-negative frames.

\paragraphbf{Lightweight optimization}
Since we only optimize a single \mbox{$D$}-dimensional vector \mbox{$\Delta$} rather than updating any VLM parameters, Q-TTA is highly computationally efficient. After optimization, $\hat{\mathbf{q}}_{\text{adapt}}$ and $\mathbf{q}_0$ are each used for multi-scale similarity fusion (Sec.~\ref{subsec:msf}). The proposals generated from the two queries are pooled and ranked jointly (Sec.~\ref{subsec:loc_head}).

\subsection{Multi-scale Similarity Fusion}
\label{subsec:msf}
Employing either frame-level or clip-level visual features in isolation yields suboptimal performance. However, because these features operate on distinct temporal grids, a direct combination of their similarity scores is non-trivial. To address this challenge, we introduce the Multi-scale Similarity Fusion (MSF) module, which aligns and integrates these complementary similarity streams into a unified and temporally structured profile.

\paragraphbf{Adapted similarity computation}
Using the refined query $\hat{\mathbf{q}}_{\text{adapt}}$, we recompute the similarities across both visual streams. Following the similarity metric defined in Eq.~\ref{eq:init_sim}, the updated frame- and clip-level similarities are:
\begin{equation}
s^{\text{frm}}_t =\langle \hat{\mathbf{q}}_{\text{adapt}}, \hat{\mathbf{v}}^{\text{frm}}_{t} \rangle; \quad s^{\text{clip}}_i=\langle \hat{\mathbf{q}}_{\text{adapt}}, \hat{\mathbf{v}}^{\text{clip}}_{i} \rangle,
\end{equation}
where $t=1,\dots,T$ and $i=1,\dots,N$.
Each stream is independently rescaled to $[0,1]$. Frame-level similarities provide fine temporal resolution but are often noisy due to missing context, whereas clip-level similarities capture broader temporal semantics but lack boundary precision. We need to integrate these complementary signals to achieve both temporal stability and localization accuracy.

\paragraphbf{Center-projection alignment}
Frame- and clip-level similarities are defined on different temporal grids. Let $f_t$ denote the timestamp of frame $t$, and $c_i$ the center timestamp of clip $i$. We project clip scores onto the frame timeline via kernel-based temporal aggregation:
\begin{equation}
\tilde{s}^{\text{clip}}_t = \sum_{i=1}^{N}
K(f_t - c_i)\, s^{\text{clip}}_i,
\quad
K(d) = \max\!\left(0,\,1 - \frac{|d|}{\delta}\right),
\end{equation}
where $K(\cdot)$ is a compact-support temporal kernel, and $d$ denotes the temporal distance. Here we adopt a triangular kernel with bandwidth $\delta$.

\paragraphbf{Score fusion}
We fuse frame- and clip-level streams as:
\begin{equation}
s^{\text{fus}}_t = (1-\alpha)\, s^{\text{frm}}_t + \alpha\, \tilde{s}^{\text{clip}}_t,
\end{equation}
where \mbox{$\alpha \in [0,1]$} is a fixed mixing hyperparameter shared across all videos. The fused score \mbox{$s^{\text{fus}}_t$} is rescaled to \mbox{$[0,1]$} and fed into the structured interval optimization (Sec.~\mbox{\ref{subsec:loc_head}}).

By integrating fine-grained semantic evidence with temporally-aware clip context, this multi-scale similarity fusion (MSF) effectively overcomes the static-feature bottleneck of prior training-free pipelines. Experiments show that MSF consistently improves single-scale baselines and complements query-side adaptation.

\begin{table*}[t]
\centering
\renewcommand{\arraystretch}{0.9}
\setlength{\tabcolsep}{8pt}
\caption{Results on Charades-STA and ActivityNet Captions. Published results use original backbones; zero-shot reproductions are grouped by embedding backbone. $\dagger$ denotes an external LLM. Within each setting block (and each backbone for reproduced results), \textbf{bold} and \underline{underlining} denote the best and second-best reported values, respectively.}
\label{tab: main results}
\begin{tabular*}{\textwidth}{@{\extracolsep{\fill}}lccccccccc}
\toprule
\multirow{2}{*}{\textbf{Method}} & \multirow{2}{*}{\textbf{Setting}} & \multicolumn{4}{c}{\textbf{Charades-STA}} & \multicolumn{4}{c}{\textbf{ActivityNet Captions}} \\
\cmidrule(lr){3-6}\cmidrule(lr){7-10}
 & & R@0.3 & R@0.5 & R@0.7 & mIoU & R@0.3 & R@0.5 & R@0.7 & mIoU \\
\midrule
\rowcolor[gray]{0.94}[0pt][\tabcolsep]
\multicolumn{10}{@{}l}{\textit{Published results with original backbones}} \\
\addlinespace[2pt]
EMB~\cite{huang2022video} & \multirow{3}{*}{fully} & \underline{72.50} & 58.33 & 39.25 & \underline{53.09} & \textbf{64.13} & \textbf{44.81} & \textbf{26.07} & \textbf{45.59} \\
EaTR~\cite{jang2023eatr} &  & - & \underline{68.47} & \underline{44.92} & - & \underline{58.18} & \underline{37.64} & - & - \\
FlashVTG~\cite{Cao_2025_WACV} &  & \textbf{80.65} & \textbf{70.32} & \textbf{49.87} & \textbf{59.61} & - & - & - & - \\
\midrule
CPL~\cite{zheng2022weakly_cvpr} & \multirow{2}{*}{weakly} & \underline{66.40} & \underline{49.24} & \underline{22.39} & - & \underline{55.73} & \underline{31.37} & - & - \\
Huang et al.~\cite{huang2023weakly} &  & \textbf{69.16} & \textbf{52.18} & \textbf{23.94} & \textbf{45.20} & \textbf{58.07} & \textbf{36.91} & - & \textbf{41.02} \\
\midrule
Luo et al.~\cite{luo2024zero} & \multirow{6}{*}{zero-shot} & 56.77 & 42.93 & 20.13 & 37.92 & 48.28 & 27.90 & 11.57 & 30.45 \\
VTG-GPT$^\dagger$~\cite{xu2024vtg} &  & 59.48 & 43.68 & 25.94 & 39.81 & 47.13 & 28.25 & 12.84 & 30.49 \\
TFVTG$^\dagger$~\cite{Zheng2024tfvtg} &  & 67.04 & \underline{49.97} & 24.32 & 44.51 & 49.34 & 27.02 & 13.39 & 34.10 \\
TAG~\cite{lee2025tag} &  & \underline{67.82} & 48.58 & \underline{26.67} & \underline{45.69} & \underline{51.88} & 28.91 & 15.07 & \underline{36.55} \\
GranAlign$^{\dagger}$~\cite{jeon2026granalign} &  & 59.10 & 39.60 & 22.70 & 38.00 & 50.30 & \textbf{34.00} & \underline{16.50} & 33.10 \\
DART-7B~\cite{dart2026} &  & \textbf{70.98} & \textbf{52.04} & \textbf{29.45} & \textbf{48.93} & \textbf{54.90} & \underline{32.14} & \textbf{18.11} & \textbf{39.89} \\
\midrule
\rowcolor[gray]{0.94}[0pt][\tabcolsep]
\multicolumn{10}{@{}l}{\textit{Backbone:}\enspace Embedding-Gemma-2-440M} \\
\addlinespace[2pt]
TFVTG~\cite{Zheng2024tfvtg} & \multirow{3}{*}{zero-shot} & 64.30 & \underline{46.56} & 22.18 & 42.25 & 46.02 & 24.32 & 12.07 & 32.06 \\
TAG~\cite{lee2025tag} &  & \underline{65.05} & 44.97 & \underline{22.31} & \underline{42.70} & \underline{49.57} & \underline{26.08} & \underline{14.16} & \underline{35.81} \\
\textbf{DSE-VTG (Ours)} &  & \textbf{69.11} & \textbf{51.42} & \textbf{28.04} & \textbf{46.70} & \textbf{54.10} & \textbf{31.32} & \textbf{15.20} & \textbf{36.91} \\
\midrule
\rowcolor[gray]{0.94}[0pt][\tabcolsep]
\multicolumn{10}{@{}l}{\textit{Backbone:}\enspace WeMM-Embedding-2B} \\
\addlinespace[2pt]
TFVTG~\cite{Zheng2024tfvtg} & \multirow{3}{*}{zero-shot} & 66.88 & \underline{49.73} & 23.44 & 44.25 & 47.66 & 25.89 & 12.86 & 32.87 \\
TAG~\cite{lee2025tag} &  & \underline{68.58} & 49.27 & \underline{25.16} & \underline{45.73} & \underline{52.92} & \underline{29.84} & \underline{15.52} & \underline{37.30} \\
\textbf{DSE-VTG (Ours)} &  & \textbf{77.18} & \textbf{59.60} & \textbf{32.80} & \textbf{52.31} & \textbf{56.60} & \textbf{33.31} & \textbf{16.38} & \textbf{38.64} \\
\midrule
\rowcolor[gray]{0.94}[0pt][\tabcolsep]
\multicolumn{10}{@{}l}{\textit{Backbone:}\enspace Qwen3-VL-Embedding-8B} \\
\addlinespace[2pt]
TFVTG~\cite{Zheng2024tfvtg} & \multirow{3}{*}{zero-shot} & 66.91 & \underline{50.03} & 22.50 & 44.03 & 46.24 & 25.19 & 12.48 & 32.12 \\
TAG~\cite{lee2025tag} &  & \underline{68.25} & 48.33 & \underline{24.89} & \underline{45.41} & \underline{52.20} & \underline{29.87} & \underline{15.13} & \underline{37.01} \\
\textbf{DSE-VTG (Ours)} &  & \textbf{75.65} & \textbf{58.58} & \textbf{32.23} & \textbf{51.30} & \textbf{54.83} & \textbf{31.97} & \textbf{16.15} & \textbf{37.93} \\
\bottomrule
\end{tabular*}
\end{table*}

\subsection{Structured Interval Optimization}
\label{subsec:loc_head}

Given the fused similarity sequence $\{s_t\}_{t=1}^{T}$ from MSF (Sec.~\ref{subsec:msf}), we localize the temporal interval that maximizes semantic contrast with respect to the query.

We first suppress low-confidence regions using adaptive thresholding, only optimizing the scores \mbox{$s_t$} that satisfy \mbox{$s_t > \mu(\mathbf{s}) - \sigma(\mathbf{s})$}, where \mbox{$\mu(\mathbf{s})$} and \mbox{$\sigma(\mathbf{s})$} are the mean and standard deviation of the fused score sequence. Before Box--Cox transformation, we add $10^{-6}$ to the retained scores. If their original minimum is non-positive, we also add its absolute value. To avoid skewed similarity distributions, we follow TAG~\mbox{\cite{lee2025tag}} to apply a Box--Cox transformation:
\begin{equation}
\bar{s}_t =
\begin{cases}
\frac{s_t^\lambda - 1}{\lambda}, & \lambda \neq 0, \\
\log(s_t), & \lambda = 0,
\end{cases}
\end{equation}
with \mbox{$\lambda$} selected via likelihood search. Then we use TCC~\mbox{\cite{lee2025tag}} to cluster frame features into \mbox{$C$} groups, which yields a partition of the timeline; discrete interval boundaries $a,b$ are defined at label transitions in the retained sequence. Candidate intervals are scored using the calibrated similarities on the retained frame sequence. Let $M$ be its length and reindex its scores as $\{\bar{s}_j\}_{j=0}^{M-1}$. For $0\leq a<b\leq M$ and $b-a<M$,
\begin{equation}
\phi(a,b)=\frac{\sum_{j=a}^{b-1}\bar{s}_j}{b-a}-\frac{\sum_{j\in\{0,\ldots,M-1\}\setminus[a,b)}\bar{s}_j}{M-(b-a)}.
\end{equation}
For the complete retained sequence, $\phi(0,M)=M^{-1}\sum_{j=0}^{M-1}\bar{s}_j$.
The candidate boundaries are mapped back to the original frame indices and timestamps.

We apply the same decoding procedure to the original and adapted query embeddings. Proposal scores $c^k$ are normalized within each branch before pooling. The pooled candidates are then ranked using overlap-weighted scores to produce the final predictions.

\vspace{2mm}
\section{Experiments}
\label{sec:exp}

\subsection{Datasets \& Evaluation Metrics}
\label{sec:data}
We evaluate on three widely used video temporal grounding benchmarks covering both in-distribution (IID) and out-of-distribution (OOD) settings.

\textbf{Charades-STA}~\cite{gao2017tall} is built upon the Charades dataset~\cite{Sigurdsson2016charades}, which focuses on indoor human activities. Its test split contains 1,334 videos with 3,720 query-moment pairs, with an average video duration of approximately 30 seconds. To further evaluate robustness under distribution shift, we additionally report results on the Novel-Composition and Novel-Word splits of Charades-CG~\cite{li2022charadesCG}, as well as the Test-OOD split of Charades-CD~\cite{charadesCD}. These splits provide complementary OOD settings for assessing generalization beyond the standard Charades-STA distribution. 

\textbf{ActivityNet Captions}~\cite{krishna2017dense} is evaluated on the val\_2 split, comprising 4,885 videos and 17,031 query-moment pairs, covering diverse open-domain activities. The videos are longer than those in Charades-STA, with an average duration of about 120 seconds, making temporal localization more challenging over extended temporal contexts.

\textbf{QVHighlights}~\cite{lei2021detecting} contains 1,550 validation queries over YouTube videos, with an average duration of approximately 150 seconds. The dataset mainly covers news and vlog content and provides annotations for both temporal moment boundaries and frame-level saliency scores.

\begin{table*}[t]
    \centering
    
    \begin{minipage}[b]{0.48\textwidth}
        \centering
        \scriptsize
        \caption{QVHighlights val results (\%). Best and second-best values are \textbf{bold} and \underline{underlined}.}
        \label{tab:qvh}
        
        \resizebox{\linewidth}{!}{%
\begin{tabular}{lccccc}
\toprule
 &  & \multicolumn{2}{c}{\textbf{Recall}} & \multicolumn{2}{c}{\textbf{mAP}} \\ \cmidrule(r{0.2em}){3-4} \cmidrule(l{0.2em}){5-6}
\multirow{-2.65}{*}{\textbf{Method}} & \multirow{-2.65}{*}{\textbf{Setting}} & @0.5 & @0.7 &@0.75& Avg. \\ \midrule
\multicolumn{1}{l|}{M-DETR \cite{lei2021detecting}} & \multicolumn{1}{c|}{} & 53.94 & 34.84 &-& 32.20 \\
\multicolumn{1}{l|}{UMT \cite{liu2022umt}} & \multicolumn{1}{c|}{} & 60.26 & 44.26 &39.90& 38.59 \\
\multicolumn{1}{l|}{QD-DETR \cite{moon2023qddetr}} & \multicolumn{1}{c|}{} & 62.68 & 46.66 &41.82& 41.22 \\
\multicolumn{1}{l|}{TR-DETR \cite{sun2024trdetr}} & \multicolumn{1}{c|}{} & 67.10 & 51.48 & 46.42 & 45.09 \\
\multicolumn{1}{l|}{$R^2$-Tuning~\cite{liu2024tuning}} & \multicolumn{1}{c|}{} & 68.71 & 52.06 & - & 47.59 \\
\multicolumn{1}{l|}{FlashVTG~\cite{Cao_2025_WACV}} & \multicolumn{1}{c|}{} & \underline{73.10} & \underline{57.29} & \underline{54.33} & \underline{52.84} \\
\multicolumn{1}{l|}{DualGround~\cite{Kang2025DualGround}} & \multicolumn{1}{c|}{\multirow{-7}{*}{fully}} & \textbf{73.48} & \textbf{58.97} & \textbf{56.35} & \textbf{53.26} \\ \midrule
\multicolumn{1}{l|}{TFVTG~\cite{Zheng2024tfvtg}} & \multicolumn{1}{c|}{} & \underline{63.68} & 38.90 & 30.14 & 32.58 \\
\multicolumn{1}{l|}{TAG~\cite{lee2025tag}} & \multicolumn{1}{c|}{} & 62.32 & \underline{44.97} & \underline{36.06} & \underline{35.75} \\
\multicolumn{1}{l|}{\cellcolor[HTML]{EFEFEF}\textbf{DSE-VTG} (Ours)} & \multicolumn{1}{c|}{\multirow{-3}{*}{\shortstack{zero-\\shot}}} & \cellcolor[HTML]{EFEFEF}\textbf{67.29} & \cellcolor[HTML]{EFEFEF}\textbf{47.16} & \cellcolor[HTML]{EFEFEF}\textbf{39.80} & \cellcolor[HTML]{EFEFEF}\textbf{38.64} \\ \bottomrule
\end{tabular}%
}
        
    \end{minipage}
    \hfill
    \begin{minipage}[b]{0.48\textwidth}
    \centering
    \scriptsize
    \caption{Results on the Charades-CD test-ood split. Best results are in \textbf{bold}; second-best are \underline{underlined}.}
    \label{tab:Charades-CD}

    \resizebox{0.85\linewidth}{!}{%
\begin{tabular}{lcccc}
\toprule &  & \multicolumn{3}{c}{\textbf{Recall}}                  \\ \cmidrule(){3-5} 
\multirow{-2.65}{*}{\textbf{Method}}     & \multirow{-2.65}{*}{\textbf{Setting}}                    & @0.3          & @0.5          & @0.7          \\ \midrule
\multicolumn{1}{l|}{SCDM \cite{yuan2019semantic}}    & \multicolumn{1}{c|}{}                        & 52.38          & 41.60          & 22.22          \\
\multicolumn{1}{l|}{MESM~\cite{liu2024mesm}}         & \multicolumn{1}{c|}{}                        & \underline{69.69}          & \underline{54.48}          & \underline{29.39}          \\
\multicolumn{1}{l|}{CICR~\cite{tang2025cicr}}        & \multicolumn{1}{c|}{}                        & \textbf{70.73} & \textbf{54.72} & 29.30          \\
\multicolumn{1}{l|}{DEMR~\cite{huang2026demr}}       & \multicolumn{1}{c|}{\multirow{-4}{*}{fully}} & 67.81          & 52.46          & \textbf{30.97} \\ \midrule
\multicolumn{1}{l|}{WSSL \cite{duan2018weakly}}      & \multicolumn{1}{c|}{weakly}                  & \textbf{35.86} & \textbf{23.67} & \textbf{8.27}  \\ \midrule
\multicolumn{1}{l|}{SPL \cite{zheng2023generating}}  & \multicolumn{1}{c|}{unsup.}                  & \textbf{62.96} & \textbf{38.25} & \textbf{15.53} \\ \midrule
\multicolumn{1}{l|}{TFVTG \cite{Zheng2024tfvtg}}     & \multicolumn{1}{c|}{}                        & 65.45          & 48.67          & 22.64          \\
\multicolumn{1}{l|}{TAG~\cite{lee2025tag}}           & \multicolumn{1}{c|}{}                        & \underline{67.70}   & \underline{50.28}          & \underline{28.47}   \\
\multicolumn{1}{l|}{\cellcolor[HTML]{EFEFEF}\textbf{DSE-VTG} (Ours)} &
  \multicolumn{1}{c|}{\multirow{-3}{*}{\shortstack{zero-\\shot}}} &
  \cellcolor[HTML]{EFEFEF}\textbf{72.21} &
  \cellcolor[HTML]{EFEFEF}\textbf{54.84} &
  \cellcolor[HTML]{EFEFEF}\textbf{31.11} \\ \bottomrule
\end{tabular}%
}
        
    \end{minipage}
\end{table*}

\begin{table}[t]
\centering
\caption{Evaluation results under OOD setting on Charades-CG.}
\label{tab:Charades-CG}
\resizebox{\linewidth}{!}{%
\begin{tabular}{lccccc}
\toprule
 &  & \multicolumn{2}{c}{\textbf{Novel-Composition}} & \multicolumn{2}{c}{\textbf{Novel-Word}} \\ \cmidrule(r{0.2em}){3-4}  \cmidrule(l{0.2em}){5-6}
\multirow{-2.65}{*}{\textbf{Method}} & \multirow{-2.65}{*}{\textbf{Setting}} & R@0.5 & mIoU & R@0.5 & mIoU \\ \midrule
\multicolumn{1}{l|}{VISA~\cite{li2022charadesCG}} & \multicolumn{1}{c|}{} & \underline{45.41} & \multicolumn{1}{c|}{\underline{42.03}} & 42.35 & 40.18  \\
\multicolumn{1}{l|}{M-DETR~\cite{lei2021detecting}} & \multicolumn{1}{c|}{} & 37.65 & \multicolumn{1}{c|}{36.17} & 43.45 & 38.37  \\
\multicolumn{1}{l|}{QD-DETR~\cite{moon2023qddetr}} & \multicolumn{1}{c|}{} & 40.62 & \multicolumn{1}{c|}{36.64} & 48.20 & 43.22  \\
\multicolumn{1}{l|}{MESM~\cite{liu2024mesm}} & \multicolumn{1}{c|}{} & 44.39 & \multicolumn{1}{c|}{39.89} & \underline{52.66} & \underline{46.38}  \\
\multicolumn{1}{l|}{SHINE~\cite{Cheng2024SHINE}} & \multicolumn{1}{c|}{\multirow{-5}{*}{fully}} & \multicolumn{1}{c}{\textbf{50.23}} & \multicolumn{1}{c|}{\textbf{44.14}} & \multicolumn{1}{c}{\textbf{55.25}} & \multicolumn{1}{c}{\textbf{48.10}} \\ \midrule
\multicolumn{1}{l|}{CPL~\cite{zheng2022weakly_cvpr}} & \multicolumn{1}{c|}{\multirow{3}{*}{weakly}} & 39.11 & \multicolumn{1}{c|}{35.53} & \underline{45.90} & -  \\
\multicolumn{1}{l|}{PPS~\cite{kim2024pps}} & \multicolumn{1}{c|}{} & \underline{40.09} & \multicolumn{1}{c|}{\underline{37.07}} & 42.01 & \underline{38.23}  \\
\multicolumn{1}{l|}{PC-Net~\cite{zhou2025pcnet}} & \multicolumn{1}{c|}{} & \textbf{41.69} & \multicolumn{1}{c|}{\textbf{38.04}} & \textbf{46.19} & \textbf{41.06} \\ \midrule
\multicolumn{1}{l|}{Luo et al.~\cite{luo2024zero}} & \multicolumn{1}{c|}{} & 40.27 & \multicolumn{1}{c|}{-} & 45.04 & -  \\
\multicolumn{1}{l|}{TFVTG~\cite{Zheng2024tfvtg}} & \multicolumn{1}{c|}{} & \underline{43.20} & \multicolumn{1}{c|}{40.43} & \underline{53.53} & 45.35  \\
\multicolumn{1}{l|}{TAG~\cite{lee2025tag}} & \multicolumn{1}{c|}{} & 43.06 & \multicolumn{1}{c|}{\underline{42.04}} & 52.81 & \underline{47.60}  \\
\multicolumn{1}{l|}{\cellcolor[HTML]{EFEFEF}\textbf{DSE-VTG} (Ours)} & \multicolumn{1}{c|}{\multirow{-4}{*}{zero-shot}} & \cellcolor[HTML]{EFEFEF}\textbf{51.77} & \multicolumn{1}{c|}{\cellcolor[HTML]{EFEFEF}\textbf{46.46}} & \cellcolor[HTML]{EFEFEF}\textbf{59.71} & \cellcolor[HTML]{EFEFEF}\textbf{50.86} \\ \bottomrule
\end{tabular}%
}%
\end{table}

\paragraphbf{Metrics} For Charades-STA and ActivityNet Captions, we adopt the same evaluation metrics as prior works~\cite{luo2024zero, Zheng2024tfvtg, lee2025tag}, including mean Intersection over Union (mIoU) and Recall@$\theta$ (R@$\theta$) at IoU thresholds $\theta \in \{0.3, 0.5, 0.7\}$. For QVHighlights, following the official evaluation protocol~\cite{lei2021detecting}, we report metrics including mAP, mAP@0.75, R@0.5, R@0.7. These metrics measure both localization accuracy and ranking quality.

\subsection{Implementation Details}
\label{sec:impl}
For video feature extraction, we employ Qwen3-VL-Embedding-8B~\cite{li2026qwen3vl-embed} as the primary vision-language backbone. Frame-level features are sampled at 3 fps. Clip-level features are extracted using a 4-second temporal sliding window with a 2-second stride and 2-second boundary segments at both ends. For the reproduced backbone comparisons in Table~\ref{tab: main results}, Embedding-Gemma-2-440M~\cite{google2026embeddinggemma2} and WeMM-Embedding-2B~\cite{zhou2026wemm} follow the same temporal sampling protocol as Qwen3-VL-Embedding-8B for both frame and clip features, while retaining their native input preprocessing. In query-level test-time adaptation, the learnable residual vector $\Delta \in \mathbb{R}^D$ is optimized via AdamW with BCE loss on pseudo-labeled frames. The default Charades-STA configuration uses up to 40 steps at learning rate $1.7\times10^{-4}$, and early stopping. Further implementation details are provided in the supplementary material.

\begin{figure*}[t]
    \centering
    \includegraphics[width=0.9\textwidth]{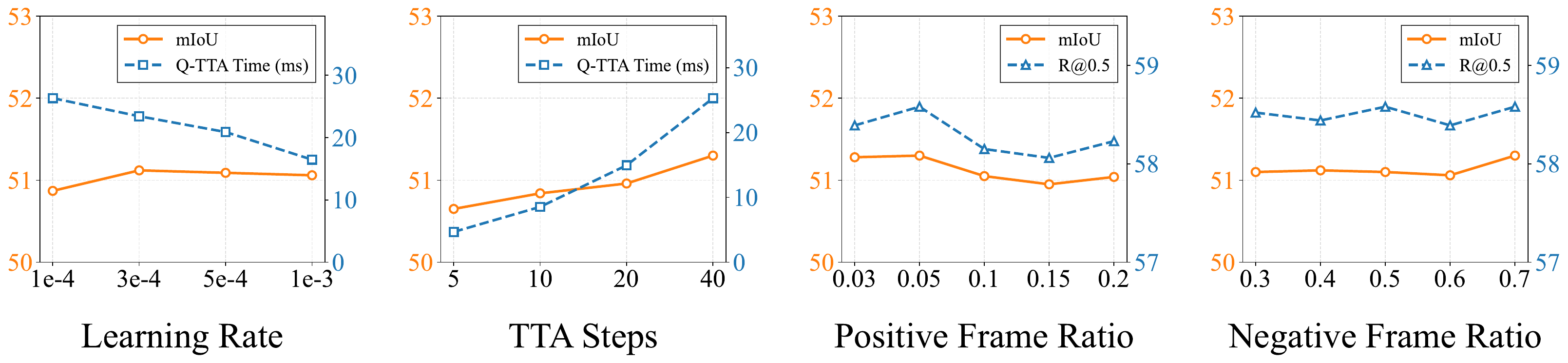}
    \caption{Hyperparameter sensitivity analysis of Q-TTA.}
    \label{fig:TTA analysis}
    \par\vspace{6pt}
    \begin{minipage}[t]{0.48\textwidth}
        \vspace{0pt}
        \centering
\scriptsize
\setlength{\tabcolsep}{6pt}
\captionof{table}{\textbf{Ablations on each component.} Best results are in \textbf{bold}; second-best are \underline{underlined}.}
\label{tab:component_ablation}
\resizebox{\linewidth}{!}{%
\begin{tabular}{ccccccc}
\toprule
\multirow{2.65}{*}{\textbf{MSF}} & \multirow{2.65}{*}{\textbf{Q-TTA}} & \multicolumn{3}{c}{\textbf{Charades-STA}} & \multicolumn{2}{c}{\textbf{QVHighlights}} \\ \cmidrule(r{0.2em}){3-5} \cmidrule(l{0.2em}){6-7} 
 &  & R@0.5 & R@0.7 & mIoU & R@0.5 & mAP \\ \midrule
 &  & 50.19 & 26.26 & 45.69 & 63.81 & 35.08 \\
 & \cmark & 52.39 & 28.17 & 47.16 & 64.77 & 36.30 \\
\cmark &  & {\ul 57.12} & {\ul 30.59} & {\ul 50.25} & {\ul 65.48} & {\ul 36.94} \\
\cmark & \cmark & \textbf{58.58} & \textbf{32.23} & \textbf{51.30} & \textbf{67.29} & \textbf{38.64} \\ \bottomrule
\end{tabular}%
}

    \end{minipage}\hfill
    \begin{minipage}[t]{0.48\textwidth}
        \vspace{0pt}
        \centering
\footnotesize
\setlength{\tabcolsep}{8pt}
\captionof{table}{\textbf{MSF ablation.} Colored dots identify \blipdot~BLIP-2~\cite{li2023blip} and \qwendot~Qwen3-VL-Embedding-8B~\cite{li2026qwen3vl-embed}.}
\label{tab:video_side_ablation}
\resizebox{\linewidth}{!}{%
\begin{tabular}{ccccccc}
\toprule
\textbf{Backbone} & \textbf{Frame} & \textbf{Clip} & & \textbf{R@0.5} & \textbf{R@0.7} & \textbf{mIoU} \\ \midrule
\blipdot   & \cmark &        & & 51.83 & {\ul 29.01} & 46.60 \\
\qwendot & \cmark &        & & {\ul 52.39} & 28.17 & {\ul 47.16} \\
\qwendot &        & \cmark & & 34.22 & 15.40 & 42.82 \\
\qwendot & \cmark & \cmark & & \textbf{58.58} & \textbf{32.23} & \textbf{51.30} \\
\bottomrule
\end{tabular}%
}

    \end{minipage}
\end{figure*}

\subsection{Experimental Results}
\label{sec:Comparison Results}
We evaluate DSE-VTG on three standard VTG benchmarks and two OOD benchmarks: Tables~\ref{tab: main results} and \ref{tab:qvh} report in-distribution results on Charades-STA, ActivityNet Captions and QVHighlights, while Tables~\ref{tab:Charades-CD} and \ref{tab:Charades-CG} report results on the three OOD splits of Charades-CD and Charades-CG under distribution shift. Owing to space constraints, additional experimental results are presented in the supplementary material.

Table~\ref{tab: main results} shows that DSE-VTG improves over TFVTG~\cite{Zheng2024tfvtg} and TAG~\cite{lee2025tag} in all reported metrics for each of the three reproduced backbones on Charades-STA and ActivityNet Captions. With Qwen3-VL-Embedding-8B~\cite{li2026qwen3vl-embed}, DSE-VTG reaches 58.58 R@0.5 and 51.30 mIoU on Charades-STA, improving over the stronger same-backbone baseline for each metric by 8.55 and 5.89 percentage points. On ActivityNet Captions, the corresponding scores are 31.97 and 37.93, exceeding TAG by 2.10 and 0.92 percentage points. The published DART result, obtained with its original backbone and inference procedure, has a higher ActivityNet mIoU of 39.89; the reproduced comparisons isolate the gains within each backbone. Additional backbone and model-size comparisons are provided in the supplementary material. The consistent gains across different backbones indicate that the effectiveness of DSE-VTG is not restricted to a particular pretrained representation.

QVHighlights~\cite{lei2021detecting} allows each query to correspond to multiple disjoint moments, requiring relevant intervals to be localized and ranked within a video. As shown in Table~\ref{tab:qvh}, DSE-VTG reaches 38.64 average mAP, exceeding the same-backbone TAG baseline~\cite{lee2025tag} by 2.89 percentage points, with gains of 2.19 and 3.74 points in R@0.7 and mAP@0.75, respectively. The joint gains in recall and mAP indicate improvements in both top-ranked localization and overall candidate retrieval. All improvements are achieved without any task-specific training or query rephrasing.

Tables~\ref{tab:Charades-CD} and \ref{tab:Charades-CG} report OOD results on Charades-CD~\cite{charadesCD} test-ood split and Charades-CG~\cite{li2022charadesCG}. DSE-VTG consistently outperforms training-free baselines under both shift types. It achieves 46.46 and 50.86 mIoU on the Novel-Composition and Novel-Word splits of Charades-CG, respectively. These results exceed TAG~\cite{lee2025tag} by 4.42 and 3.26 percentage points. R@0.5 also improves by 8.71 points on Novel-Composition. On Charades-CD, DSE-VTG further reaches 72.21 and 54.84 in R@0.3 and R@0.5, outperforming TAG by 4.51 and 4.56. These results show improved grounding performance across the evaluated OOD splits without task-specific training.

\subsection{Ablation Studies}
\label{sec:ablation}
We conduct extensive ablation studies and further analysis on Charades-STA~\cite{gao2017tall} and QVHighlights~\cite{lei2021detecting}. Additional analyses are provided in the supplementary material.

\paragraphbf{Ablation of the Dual-side Design}
Table~\ref{tab:component_ablation} isolates the contributions of the two proposed components. On Charades-STA, Q-TTA and MSF individually improve mIoU from 45.69 to 47.16 and 50.25, respectively, while their combination reaches 51.30. The same trend holds on QVHighlights, where the full model improves mAP from 35.08 to 38.64. In these component comparisons, combining both modules yields larger gains than either alone.

\paragraphbf{Video-side Multi-scale Fusion}
Table~\ref{tab:video_side_ablation} evaluates the contribution of MSF by comparing frame-only, clip-only, and fused visual inputs with Q-TTA enabled. With Qwen3-VL-Embedding-8B, the two single-scale inputs achieve 47.16 and 42.82 mIoU, respectively. Integrating them through MSF raises mIoU to 51.30 and R@0.7 to 32.23, exceeding the stronger frame-only configuration by 4.14 and 4.06 percentage points. Thus, clip features have complementary value even though their standalone localization accuracy is lower. These gains show that MSF effectively combines complementary frame and clip information.

\paragraphbf{Computational cost} Video encoding is a shared preprocessing step in feature-based training-free VTG methods~\cite{Zheng2024tfvtg,lee2025tag}. On an NVIDIA RTX 4090, extracting frame- and clip-level features takes 4.85\,s and 3.12\,s, respectively, but is performed only once per video; the resulting features are cached and reused across all queries associated with that video. Once indexed, DSE-VTG processes each query in 69\,ms end to end, including 24\,ms for text encoding, 25\,ms for Q-TTA, 1\,ms for MSF, and 19\,ms for structured interval optimization. Dual-scale features add 0.44 GB of peak memory relative to single-scale.

\paragraphbf{Hyperparameter Sensitivity and Efficiency of Q-TTA} We evaluate the hyperparameter sensitivity and efficiency of Q-TTA on Charades-STA using Qwen3-VL-Embedding-8B. Figure~\ref{fig:TTA analysis} shows modest accuracy changes across the displayed learning rates and sampling ratios. Increasing the allowed adaptation steps improves mIoU in the shown sweep and increases adaptation time. The lower timings at larger learning rates are consistent with the early-stopping rule, although the figure alone does not isolate its contribution. The 40-step setting takes approximately 25\,ms. The two rightmost panels of Fig.~\mbox{\ref{fig:TTA analysis}} show robustness to the positive and negative frame ratios.

\section{Conclusion}
\label{sec:concl}
We presented DSE-VTG, a training-free framework for video temporal grounding that combines multi-scale temporal evidence with video-specific query adaptation. Extensive experiments on three standard and two OOD benchmarks demonstrate consistent performance gains over strong baselines, with improvements observed across different backbones. DSE-VTG demonstrates the value of jointly refining temporal evidence and query representations for effective grounding without task-specific training.

{
    \small
    \bibliographystyle{ieeenat_fullname}
    \bibliography{main}
}

\end{document}